\documentclass[letterpaper, 10 pt, conference]{ieeeconf} 
\IEEEoverridecommandlockouts
\usepackage{cite}
\usepackage{amsmath,amssymb,amsfonts}
\usepackage[ruled, vlined, linesnumbered]{algorithm2e}
\usepackage{graphicx}
\usepackage{textcomp}
\usepackage{xcolor}
\usepackage{siunitx}
\usepackage{amssymb}
\def\BibTeX{{\rm B\kern-.05em{\sc i\kern-.025em b}\kern-.08em
    T\kern-.1667em\lower.7ex\hbox{E}\kern-.125emX}}
\usepackage{diagbox}
\usepackage{multirow}
\begin{document}

\title{Efficient RRT*-based Safety-Constrained Motion Planning for Continuum Robots in Dynamic Environments}

    \author{Peiyu Luo\textsuperscript{†}, Shilong Yao\textsuperscript{†}, Yiyao Yue, Jiankun Wang\textsuperscript{*}, \IEEEmembership{Senior Member, IEEE},\\
    Hong Yan, \IEEEmembership{Fellow, IEEE},
    Max Q.-H. Meng\textsuperscript{*}, \IEEEmembership{Fellow, IEEE}
    \thanks{This work was partly supported by Shenzhen Key Laboratory of Robotics Perception and Intelligence (ZDSYS20200810171800001) and the Hong Kong Innovation and Technology Commission (InnoHK Project CIMDA).}
    \thanks{P. Luo, S. Yao, Y. Yue, J. Wang and M. Meng are with Shenzhen Key Laboratory of Robotics Perception and Intelligence and the Department of Electronic and Electrical Engineering, Southern University of Science and Technology, Shenzhen, China. J. Wang is also with the Jiaxing Research Institute, Southern University of Science and Technology, Jiaxing, China.}
    \thanks{S. Yao and H. Yan are with the Department of Electrical Engineering, City University of Hong Kong, Hong Kong, China.}
    \thanks{$\dagger$Peiyu Luo and Shilong Yao are co-first authors.}
    \thanks{$*$Corresponding authors: Jiankun Wang, Max Q.-H. Meng. e-mail: wangjk@sustech.edu.cn, max.meng@ieee.org.}
    }    

\maketitle

\begin{abstract}
   Continuum robots, characterized by their high flexibility and infinite degrees of freedom (DoFs), have gained prominence in applications such as minimally invasive surgery and hazardous environment exploration. However, the intrinsic complexity of continuum robots requires a significant amount of time for their motion planning, posing a hurdle to their practical implementation. To tackle these challenges, efficient motion planning methods such as Rapidly Exploring Random Trees (RRT) and its variant, RRT*, have been employed. This paper introduces a unique RRT*-based motion control method tailored for continuum robots. Our approach embeds safety constraints derived from the robots' posture states, facilitating autonomous navigation and obstacle avoidance in rapidly changing environments. Simulation results show efficient trajectory planning amidst multiple dynamic obstacles and provide a robust performance evaluation based on the generated postures. Finally, preliminary tests were conducted on a two-segment cable-driven continuum robot prototype, confirming the effectiveness of the proposed planning approach. This method is versatile and can be adapted and deployed for various types of continuum robots through parameter adjustments.
    
\end{abstract}


\section{Introduction}
\PARstart{C}{ontinuum} robots are highly flexible and versatile, with extensive applications in industrial production, medical surgery\cite{2023Yao}, and rescue missions\cite{2021Mohammed}. 
Unlike traditional robots, they excel in dynamic environments and complex shapes due to their ability to bend, stretch, twist, and deform. This unique characteristic allows them to perform a wide range of movements and operations, making them adaptable and capable of exploring complex spaces\cite{2010Webster}. With increased degrees of freedom, continuum robots offer flexibility and adaptability in performing tasks like space exploring\cite{2015Jessica}. Their ability to adjust their complex shapes and navigate dynamic environments makes them ideal for handling uncertainties and obstacles, enabling them to navigate challenges and adapt to different working conditions.

\begin{figure}[h]
    \centering
    \includegraphics[width=0.4\textwidth]{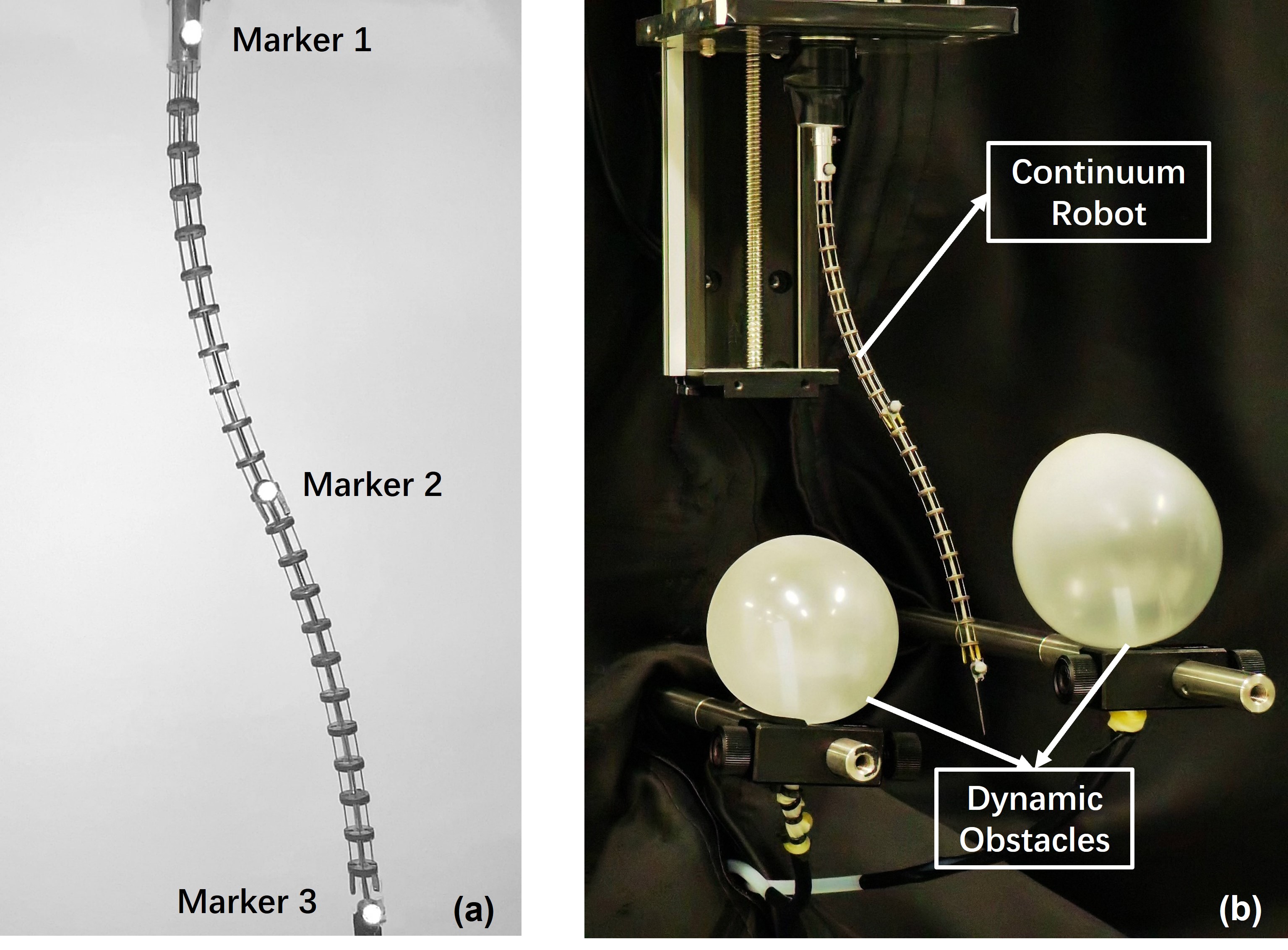}
    \caption{The robot configuration. (a). Two-section continuum robot. (b). Real-world dynamic environment implementation.}
    \label{fig:Intro}
\end{figure}
\vspace*{-2em}

Motion planning algorithms are crucial for robot navigation. Specifically, they consider map information, environmental features, velocity limitations, and dynamic obstacles. Predictive motion planning combines sampling-based algorithms like Rapidly Exploring Random Trees (RRT) and its advanced version, RRT*, known for their simplicity, adaptability, and ability to handle dynamic changes. These algorithms have widespread applications in various fields, including robotics, autonomous vehicles, and industrial automation, improving safety and efficiency in complex settings.
In the last decades, researchers have explored planning approaches for continuum robotics to be compliant and smooth. Different approaches have been developed to avoid known static obstacles\cite{2010Xiao, 2018Seleem}, while sensor-based approaches have tackled unknown static obstacles for continuum robots\cite{1992Reznik, 2009Chen}. Potential field-based planning has shown good performance, but its implementation has been limited to restricted settings\cite{2016Ataka}. The temporal graph method can also solve the problem, but it requires more time\cite{2021Meng, 2019Deng, 2022Meng}. Learning-based algorithm methods have shown promise for real-time planning\cite{2020Wang}. However, navigating dynamic environments presents significant challenges for continuum robots. Although sampling-based approaches (such as RRT* and PRM) are widely used in robotic motion planning, there have been few attempts to apply them to continuum arms. Due to their ability to real-time adapt to obstacles, continuum robots require robust sensing, decision-making, and control capabilities to ensure safe and efficient navigation.

To address these challenges, we propose a novel RRT*-based motion planner for a 6mm wide continuum robot (Fig. \ref{fig:Intro}) operating in dynamic environments . Our planner utilizes the optimization-Jacobian-based method with the piecewise constant curvature (PCC) model to smooth the trajectory. We evaluate its performance with dynamic obstacles and demonstrate successful trajectory tracking and collision avoidance. Our major contributions are as follows:

\begin{itemize}
\item We pioneer applying the RRT*-based method for motion planning in dynamic environments using continuum robots. Our approach integrates optimization constraints from the Jacobian matrix with the inherent characteristics of soft robot models, emphasizing geometrically-based safety constraints. The planner demonstrated effective control in environments with both static and dynamic challenges.

\item We further adapted the planner for scenarios with more dynamic obstacles, evaluating its performance in simulations. Our method notably reduced computation time and achieved a higher success rate compared to \cite{2021Meng}. Physical experiments confirmed its feasibility for real-time control.
\end{itemize}

\vspace{-1em}
\begin{figure}[h]
    \centering
    \includegraphics[width=0.4\textwidth]{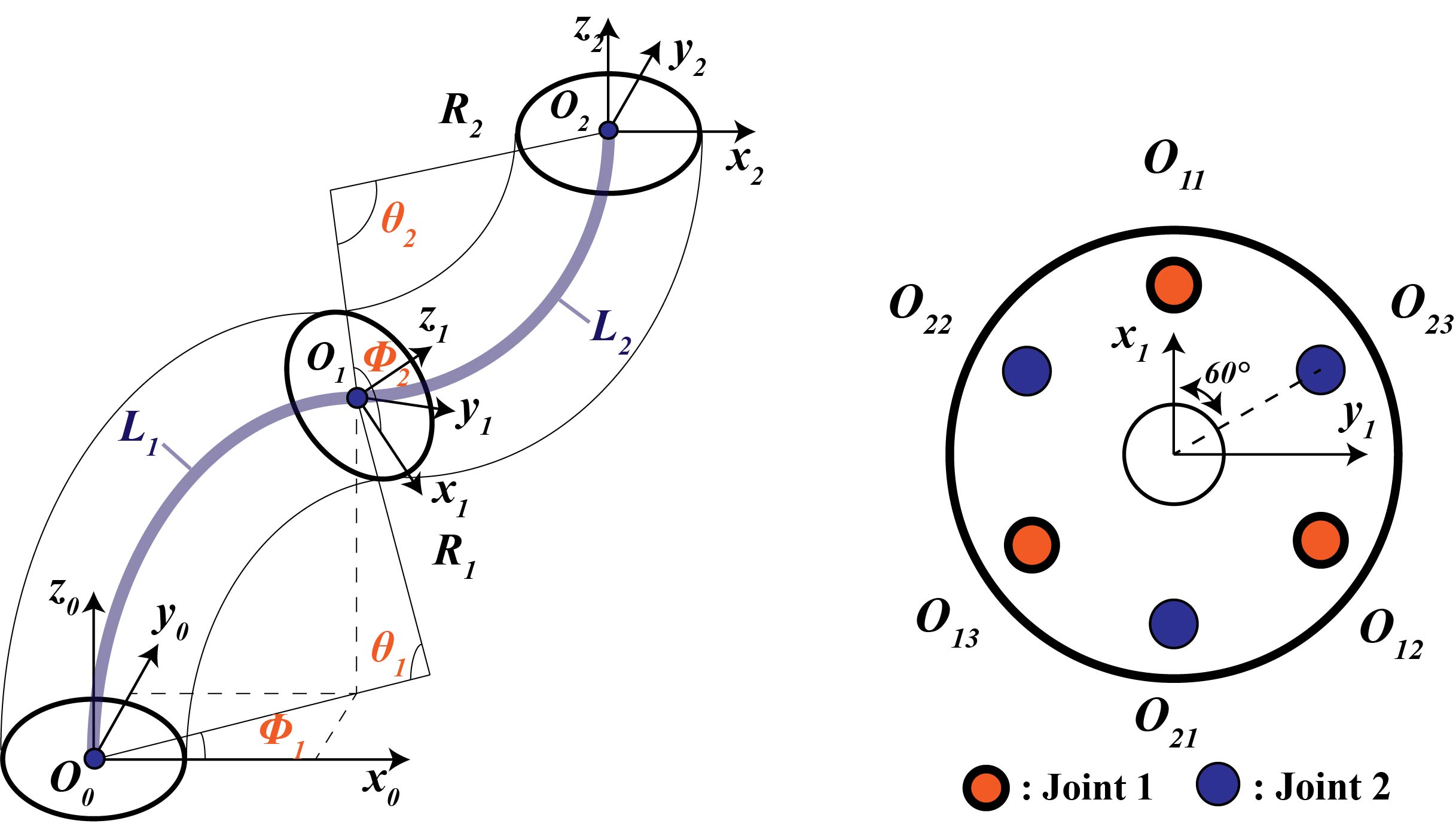}
    \caption{Continuum robot model diagram and cable arrangement.}
    \label{fig:CR_Model}
\end{figure}

\section{System Model}
Our continuum robot consists of a 0.6mm NiTi wire as a flexible backbone along the neutral axis, 6mm-width discs evenly distributed at equidistant intervals along the backbone, and driving cables. The discs have uniformly distributed holes for the drive cables. Therefore, bending the robot in various directions can be achieved by controlling the extension and contraction of the drive cables. The $k^{th}$ segment (where $k\leq n$) can be characterized by the PCC model\cite{2020Gonthina} parameters 
$(\theta_k, \phi_k, R_k)$ depicted in the Fig. \ref{fig:CR_Model}, where $\theta_k \in [ 0, \pi ]$, $\phi_k \in [ -\pi, \pi ]$ and $R_k = L_k / \theta_k$. The $\theta_k$ serves as the bending angle, $\phi_k$ denotes the bending direction angle and $R_k$ represents the bending radius of the $k^{th}$ segment. The end position of 
the $k^{th}$ segment can be formulated by these parameters as:

\begin{equation}
    p_{k-1}^{k} = \frac{L_k}{\theta_k}
    \begin{bmatrix}
    (1-\cos{\theta_k}) \cos{\phi_k} \\
    (1-\cos{\theta_k}) \sin{\phi_k}  \\
    \sin{\theta_k}
    \end{bmatrix}
    \end{equation}

Based on the model in Fig. \ref{fig:CR_Model} and \cite{2010Webster}, the parameters $\theta_k$ and $\phi_k$ of the robot can be derived as follows:

\vspace{-0.2cm}
\begin{equation}
    \theta_k(\boldsymbol{q}_k) = \dfrac{2\sqrt{q_{k,1}^2+q_{k,2}^2+q_{k,3}^2-q_{k,1}q_{k,2}-q_{k,2}q_{k,3}-q_{k,1}q_{k,3}}}{3r}
\end{equation}\vspace{-0.5cm}

\begin{equation}
    \phi_k(\boldsymbol{q}_k) = \operatorname{atan2}(\sqrt{3}(q_{k,2}-q_{k,3}), q_{k,2}+q_{k,3}-2q_{k,1})
\end{equation}
where $r$ refers to the radius between the center of the disc and the cable channel. $q_{k,i}$ is defined as the length of the $i^{th}$ cable, and
$\boldsymbol{q}_k$ represents a vector composed of $q_{k,i}$ ($i \in \{1, 2, 3\}$). From the equation above, the $R$ can be determined by the segment length and the bending angle $\theta_k$: $R_k = L_k/\theta_k(\boldsymbol{q}_k)$. 

For each robot segment, a minimum of $3$ drive cables is required for motion control. By adjusting the length of the drive cables, the robot segment can be bent in the desired direction. Based on the illustration and geometric relationships provided in Fig. \ref{fig:CR_Model}, the relationship between the length 
of the drive cables and the robot's parameters can be derived as,
\begin{equation}
    q_{k,i} = L_k - r\theta_k \cos{(\phi_k+(i-1)\xi)}
\end{equation}
where $L_k$ refers to the length of $k^{th}$ segment's backbone and $i \in \{1,2,3\}$. $\xi$ represents the angular spacing between every two drive cables. Since
the three cables are evenly distributed, $\xi$ equals to $\frac{2\pi}{3}$. By differentiating both sides of eq.(4), the instantaneous model
of the $k^{th}$ segment and the Jacobian matrix that connects the robot parameters with the change in the length of drive cables can be derived as follows,

\begin{equation}
    \dot{\boldsymbol{q}_k} = \boldsymbol{J}_k \dot{\boldsymbol{\Phi}_k}
\end{equation}\vspace{-1cm}

\begin{equation}
    \boldsymbol{J}_k = 
    \begin{bmatrix}
    -r\cos{\phi_k} && r\theta_k\sin{\phi_k} \\
    -r\cos{(\phi_k+\xi)} && r\theta_k\sin{(\phi_k+\xi)} \\
    -r\cos{(\phi_k+2\xi)} && r\theta_k\sin{(\phi_k+2\xi)} \\
    \end{bmatrix}
    \end{equation}
where $\boldsymbol{J}_k$ represents the Jacobian matrix of the $k^{th}$ segment, and $\boldsymbol{\Phi}_k=(\theta_k, \phi_k)^T \in \mathbb{R}^2$ denotes the parameters of $k^{th}$ segment.
The relationships between robotic joints can be obtained using rotation matrices. As shown in Fig. \ref{fig:CR_Model}, the endpoint of the $(k-1)^{th}$ segment serves as the base for the $k^{th}$ segment. Consequently, the tip position of the $k^{th}$ segment can be represented using a sequential multiplication of rotation matrices. The transformation matrix can be derived from tip frame of the $(k-1)^{th}$ segment and the $k^{th}$ segment's robotic parameters:
\begin{equation}
    \boldsymbol{R}_{k-1}^{k} = \text{Rot}(\overline{z}_{k-1}, \phi_k) \cdot \text{Rot}(\overline{y}_{k-1}, \theta_k) \cdot \text{Rot}(\overline{z}_{k}, -\phi_k) 
\end{equation}

For multi-segment soft robots, there exists interdependence between adjacent segments. To obtain more precise solutions, the introduction of force models for the soft segments is imperative. The bending deformation of a single segment is proportionally governed by $\theta_k$, segment length $L_k$, and bending stiffness coefficient determined by the materials denoted as $K_B$. Assuming a resultant moment of force denoted as $\tau$ acting on the tip of soft robot, then $\theta_k$ can be formulated as \cite{2020Gonthina}:
\begin{equation}
    \theta_k(\boldsymbol{q}_k) = \alpha \dfrac{L_k \tau}{K_B} = \dfrac{M_k}{K_B}
\end{equation}
where $\alpha$ is the proportional coefficient.

The bending of the $(k-1)^{th}$ segments is influenced by both their intrinsic control parameters, denoted as $\boldsymbol{q}_{k-1}$, and the successive parameters represented by $\boldsymbol{q}_{k}$. Consequently, the torque exerted on the $(k-1)^{th}$ segments is subject to the influence of both their individual characteristics and the torque generated by the $k^{th}$ segment. Assuming that $\overline{M}_{n-1}$
represents the resultant moment applied to the $(k-1)^{th}$ segment. The interrelationship among coupled segments is described as follows,
\begin{equation}
    \overline{M}_{k-1} = M_{k-1} + \overline{M}_{k}
\end{equation}
where $\overline{M}_k \equiv M_k$ when $k=n$.

The robot structure has negligible elongation along its primary axis, thus $L_k$ can be considered invariant. Substituting eq.(9) into the triangular configuration equation in \cite{2020Gonthina}, the explicit relationship can be analytically deduced as:
\begin{equation}
    \overline{\theta}_{k-1} = \dfrac{M_{k-1}^2}{K_B} + K_B \theta_k^2 + 2 M_{k-1} \theta_k \cos{(\phi_{k-1}-\phi_{k}+\pi)^{\frac{1}{2}}}
\end{equation}
\begin{equation} 
    \begin{split}
    \overline{\phi}_{k-1} = \dfrac{\pi}{2} + \operatorname{atan2} \{M_{k-1} \sin{\phi_{k-1}}+K_B\theta_k \sin{(\phi_k-\pi)},\\
    M_{k-1} \cos{\phi_{k-1}}+K_B\theta_k \cos{(\phi_k-\pi)} \}
    \end{split}
\end{equation}\vspace{-0.2cm}

From eq.(10) and eq.(11), the coupled configuration can be derived. This allows for the deduction of segment-specific configuration parameters from cable actuation variations. As a result, the robot's geometric parameters are inferred, facilitating obstacle avoidance at the geometric level.

\section{Methodology}
In this section, we present our proposed planning method from two aspects. Initially, we utilize the optimization-Jacobian-based method to achieve robust control for the robot. We ensure safe distances between the robot and obstacles by constraining its Jacobian matrix, facilitating effective geometric obstacle avoidance. Besides, we propose a control algorithm that combines the optimization-Jacobian-based method with RRT* to improve the adaptability and efficiency of continuum robots in dynamic environments.

\subsection{Optimization-Jacobian-Based Control Method}
The investigation within the following methods are models based on two-segment robots. Our two-segment continuum robot comprises two sets of six actuation cables and a z-axis translational input serving as the actuation. Each set of cables is positioned on the circular disk with a 180-degree phase difference. The drive vectors are to be represented as:
\begin{equation}
    \boldsymbol{q} = [q_1, q_2, q_3, q_4, q_5, q_6, z]^T
\end{equation}

The position error at the tip can be represented by $p_d=x_d-x_c$. Thereby transforming the positional error at the tip into errors within the robot's configuration space\cite{2011Bajo},
\begin{equation}
    \boldsymbol{e_{\Phi}} = \Phi_d - \Phi_c
\end{equation}
substituting eq.(5) into the derivative of eq.(13) yields,
\begin{equation}
    \boldsymbol{\dot{e_{\Phi}}} = \dot{\Phi_d} - \rho \dot{\boldsymbol{J}_{\Phi}^\dagger \dot{\boldsymbol{q}}}
\end{equation}
where $\rho \in \mathbb{R}$ is determined empirically. By rearranging eq.(14), a linear proportional controller can be obtained \cite{2022LaiTmech},
\begin{equation}
    \boldsymbol{\dot{q}} = \boldsymbol{J}_{\Phi}^\dagger(\dot{\Phi}_d + \boldsymbol{K}_P \boldsymbol{e_{\Phi}})
\end{equation}
where $\boldsymbol{K}_P$ is a symmetric positive definite matrix\cite{2022LaiRAL}. When reaching the desired position, the robot ceases motion. Thus, $\Phi_d=0$, and eq.(15) can be modified as follows,
\begin{equation}
    \boldsymbol{\dot{q}} = \boldsymbol{J}_{\Phi}^\dagger\boldsymbol{K}_P \boldsymbol{e_{\Phi}}
\end{equation}

$\boldsymbol{\dot{q}}$ can be approximated as $\Delta \boldsymbol{q}$ in a extremely short period. By adjusting $\Delta \boldsymbol{q}$ to narrow the gap with the desired configurations, the tip position is brought closer to the desired position. To mitigate the infinite solutions arising from the system redundancy, the damped least squares (DLS) pseudo-inverse method is introduced\cite{2022LaiRAL}. The optimization problem can be addressed using damped convex least squares,
\begin{equation}
    \underset{\Delta \boldsymbol{q}}{\arg\min} (\|\boldsymbol{J}_{\Phi} \Delta \boldsymbol{q}-\boldsymbol{K}_P \boldsymbol{e_{\Phi}}\|^2_2+\lambda^2\|\Delta \boldsymbol{q}\|^2_2)
\end{equation}
where $\lambda$ represents a positive damping constant. This model constructs a quadratic equation to minimize the error between the optimized robot parameters and the desired configurations, aiming for minimal disparity. The resultant tip position approximates the desired location. To avert potential damage or singular solutions caused by excessive amplitudes during actuation, the second constraint term is incorporated to ensure smooth movement. When addressing complex environments, additional constraints are considered during the planning process to accommodate constraints from both the environment and the mechanical model\cite{2022LaiRAL}.

\begin{equation}
    \underset{\Delta \boldsymbol{q}}{\arg\min} \ \mu^2 \| \frac{R_{obs}+D_{sf}}{\min \|L(\Delta \boldsymbol{q})-C_{obs} \|} \|_2^2
\end{equation}
\vspace{-0.2cm}

Eq.(18) represents an optimization function employed for obstacle avoidance, where $R_{obs}$ denotes the radius of obstacles, $C_{obs}$ represents the center of obstacles and $D_{sf}$ signifies the safety threshold distance between the robot and obstacles. $L(\cdot)$ is a function enabling the discrete representation of the robot's spatial position. $u$ corresponds to the coefficient governing the optimization terms. Eq.(18) serves as the core function for robot safety constraints, used to perform collision detection during the planning process.

\begin{algorithm}
  \LinesNumbered
  \caption{\mbox{RRT*-based Safety-Constrained Algorithm.}}
  \label{alg:alg1}
  \KwIn{Initial point $P_{init}$; target point $P_{goal}$; obstacles $\mathcal{O}$; maximum number of iterations $m$}
  \KwOut{A sequence of reachable points as path $\mathcal{P}$ and desired corresponding configurations $\Psi$}
  
  Build tree $\mathcal{T}$ and search space $\mathcal{X}$\;
  
  Add initial point $P_{init}$ to $\mathcal{T}$\;
 
  \While{number of iterations is less than $m$}{
    Generate a new point $P_{new}$ in $\mathcal{X}$\;
    \If{$P_{new}$ is valid and obstacle-free}{
        Find the nearest point $n$ to $P_{new}$\;
      \If{steer($P_{new}$, $n$) is valid and obstacle-free during the steer process}{
        Calculate a configuration for $P_{new}$ using optimization method\;
        
        Add $P_{new}$ into $\mathcal{T}$ and rewire\;
        
        Record as one valid iteration\;
      }
      \Else{
        Generate another point $P_{new}'$ and repeat the above process\;
        }
      }
      Check to see if $P_{goal}$ is in the tree\;
  }
  Generate path $\mathcal{P}$ and configurations $\Psi$\;
  
  \Return $\mathcal{P}$ and $\Psi$\;
\end{algorithm}

\subsection{RRT*-Based Safety-Constrained Algorithm}
This subsection introduces a motion planning control algorithm based on RRT* that incorporates the constraints mentioned in the first part. The algorithm \ref{alg:alg1} takes the initial and goal positions, along with the parameters of the robot and the obstacles. The initial step involves constructing a tree denoted as $\mathcal{T}$ and creating a search space $\mathcal{X}$ based on obstacles and the search range. The starting point $P_{init}$ is inserted into the tree. The iterative point search begins at line 3. A point $P_n$ is randomly sampled in each iteration. If $P_n$ fulfills the criteria of being within an obstacle-free region and remaining within the boundaries of the sampling space, the nearest point $n$ to $P_{new}$ within the tree is determined. Utilizing an optimization-Jacobian-based method within the $steer(\cdot)$ function, an optimal solution is sought and then subjected to an obstacle-free check. Assuming that the robot can avoid collisions during the steering process from $n$ to $P_{new}$, the computed configuration is recorded. Subsequently, the point $P_{new}$ is inserted into the tree and marked as a valid sampling instance (lines 8-10). In non-valid sampling instances, a random sample point $P_{new}'$ is drawn, and the aforementioned validation procedure is reiterated. Finally, the presence of $P_{goal}$ within the tree is checked; if $P_{goal}$ is reached, the algorithm returns the path $\mathcal{P}$ connecting the start and goal points with corresponding configurations $\Psi$.

\section{Simulation}
In this section, we validate our proposed RRT*-based safety-constrained method. We demonstrate its efficacy in motion planning and obstacles avoidance for both static and dynamic environments. Our approach employs a two-segment Cable-Driven Soft (Continuum) Robot (CDSR) model designed to account for coupling effects. The simulations were implemented using Python, incorporating parameters derived from our prototype. All simulations were executed on a laptop with an Intel Core i7-12700H CPU @2.30GHz 32GB RAM.

\subsection{Motion Planning in Static and Dynamic Environment}
\begin{figure}[h]
    \centering
    \includegraphics[width=0.5\textwidth]{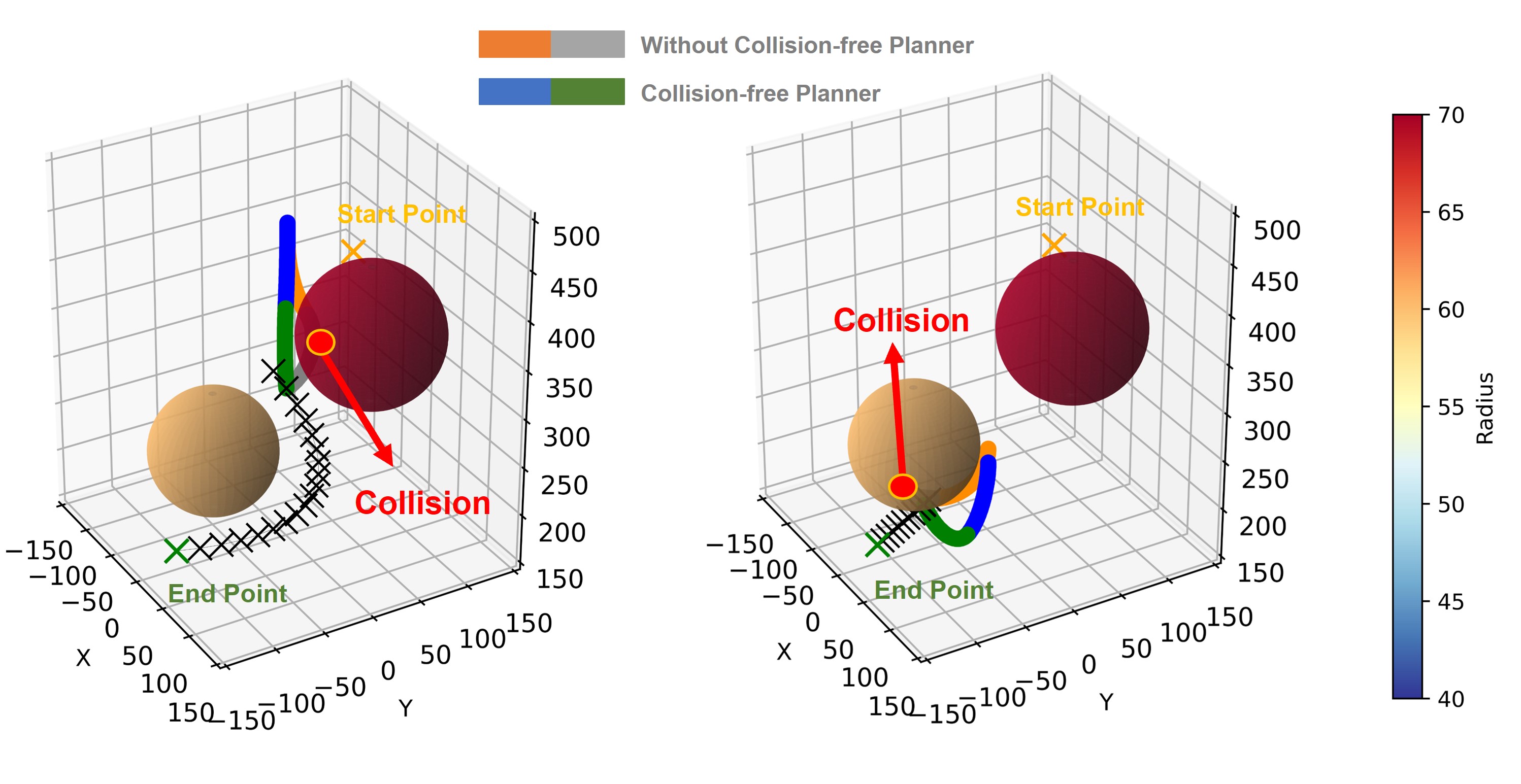}
    \caption{Simulation 1: Static environment motion planning comparison. The black intersections represent the planned path, indicating a collision-free route from the starting point to the endpoint, while the yellow and green intersections represent the task's start and end points.}
    \label{fig:simulation1}
\end{figure}

The first part involves simulating motion planning in a static environment with obstacles. The simulation results are shown in Fig. \ref{fig:simulation1}. The Orange-Grey and Blue-Green planners show the robot's posture without and with obstacle considerations, respectively. The search space is a 300mm$\times$300mm$\times$500mm rectangular region centered around the robot's base axis. Two spherical obstacles are placed at $[0, 90, 330]^T$ and $[-10, -70, 250]^T$ with radius of 65mm and 55mm. The start and destination points are $[-50, 100, 390]^T$ and $[10, -120, 170]^T$, respectively. The objective of the motion planning is to ensure collision-free of the robot's body with the obstacles. Empirically, we set the damping coefficient for constraint terms to 0.3 and the obstacle-free coefficient to 20. The approach aligns with CDSR's kinematics by sampling positions and calculating optimized inverse kinematics (IK) solutions. The single computation of the IK solution takes approximately 0.47s. This iterative process generates a collision-free path, followed by spline curve interpolation for smooth execution. The CDSR shall complete tracking the path and avoiding collisions with obstacles. The time required for finding a solution is about 2.41s on average. 

\begin{figure*}[t]
    \centering
    \begin{minipage}[t]{0.49\textwidth}
        \centering
        \rule{\linewidth}{0.5pt}
        \text{(a) With Collision-free Planner}
        \rule{\linewidth}{0.5pt}
        \includegraphics[width=1.0\linewidth]{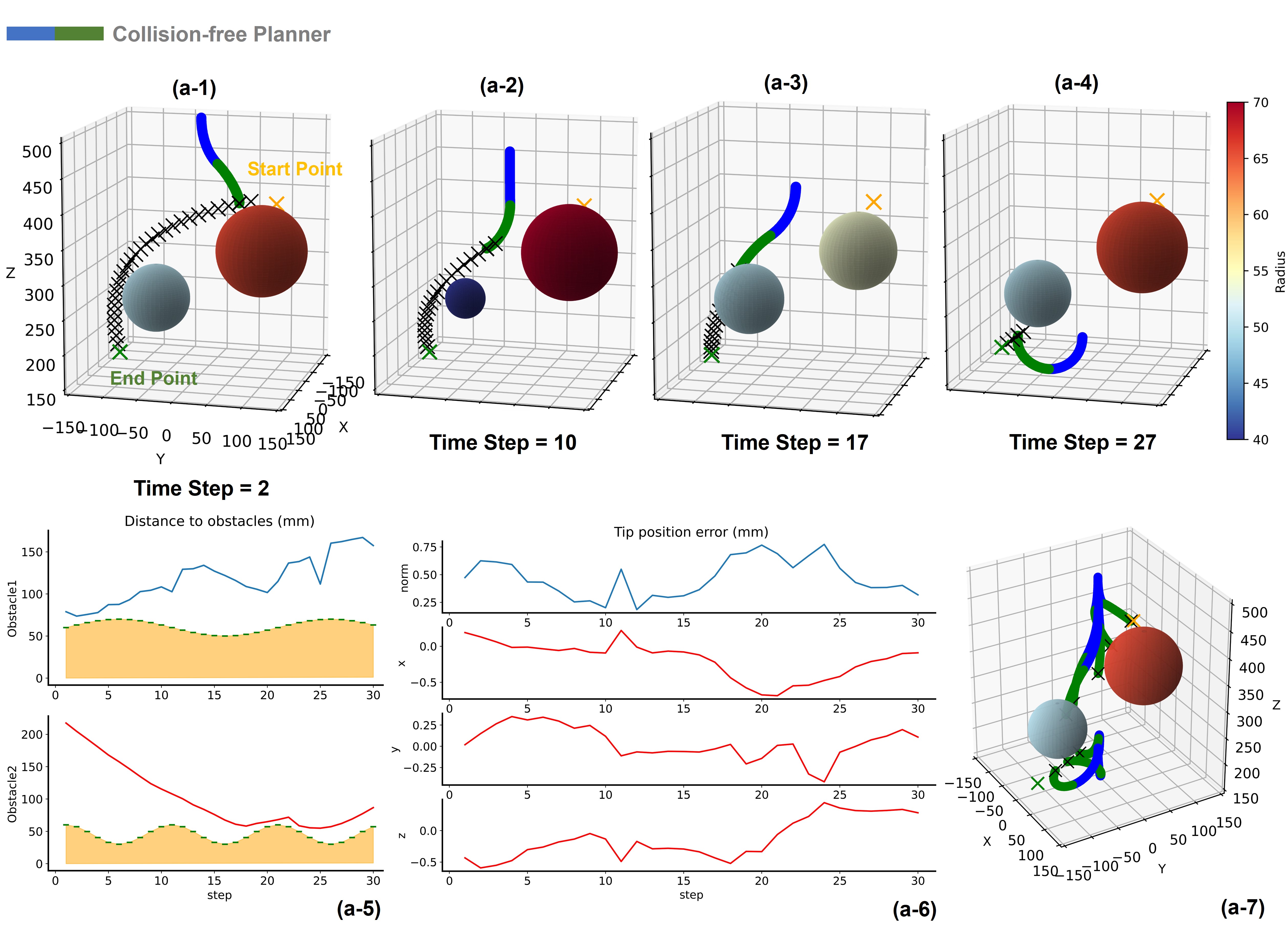}
        \rule{\linewidth}{0.5pt}
    \end{minipage}
    \hfill
    \begin{minipage}[t]{0.49\textwidth}
        \centering
        \rule{\linewidth}{0.5pt}
        \text{(b) Without Collision-free Planner}
        \rule{\linewidth}{0.5pt}
        \includegraphics[width=1.0\linewidth]{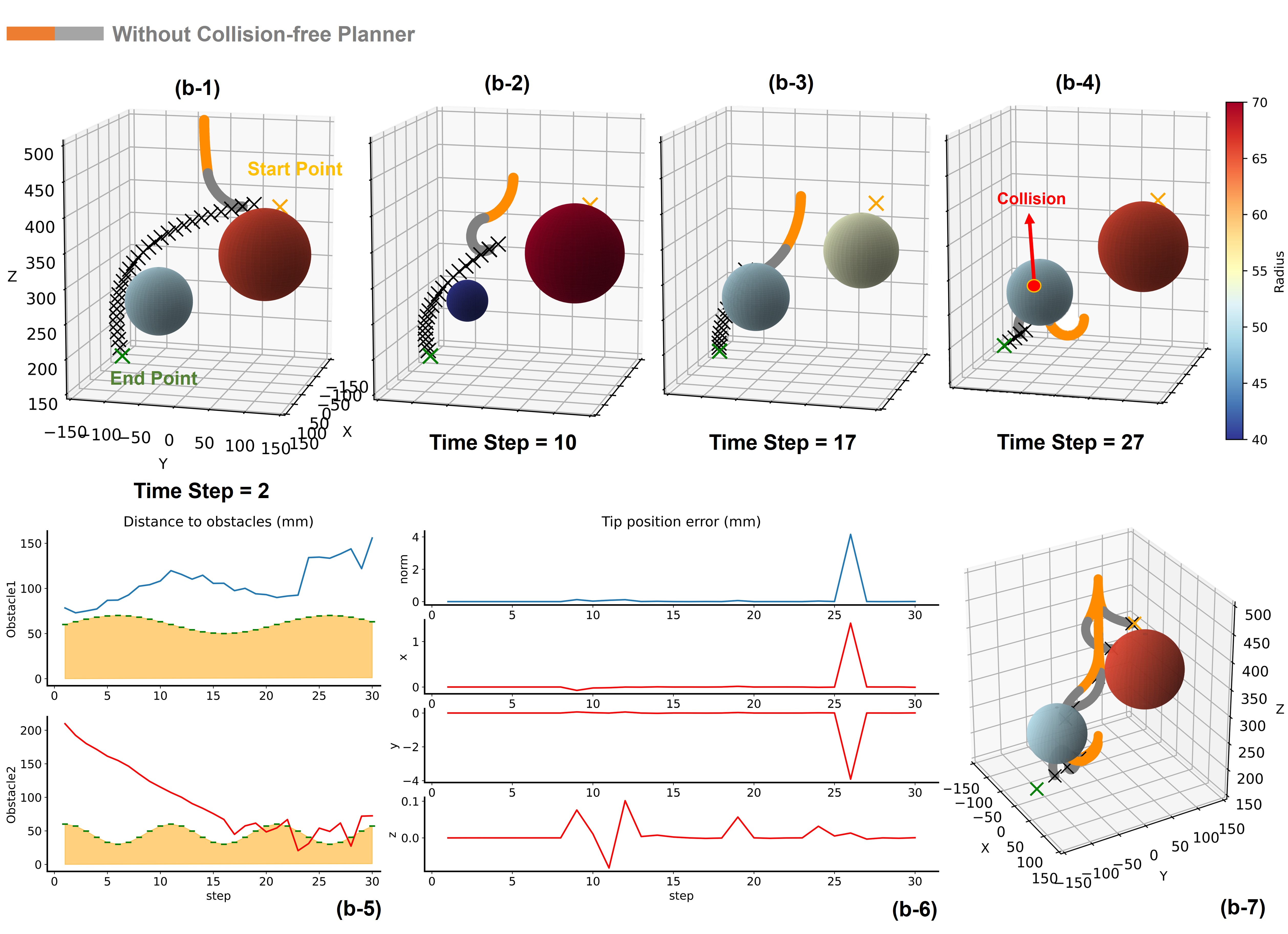}
        \rule{\linewidth}{0.5pt}
    \end{minipage}
    \caption{Simulation 2: Dynamic environment motion planning. (a). By employing appropriate motion planning strategies, the CDSR can avoid collisions while maintaining satisfactory path-tracking accuracy. (b). During the path exploration process, the CDSR may collide with obstacles, which occur within the red point. All conditions remain the same, except for the inclusion or exclusion of obstacle-free motion planning.}
    \label{fig:simulation2}
\end{figure*}
\vspace{-0.2cm}

It can be observed that when the robot does not account for obstacles (Orange-Grey planner), the tip follows the interpolated points, but the robot's body partially intersects with the obstacles (see Fig. \ref{fig:simulation1}). The Blue-Green planner illustrates the situation after incorporating obstacle avoidance. The solver provides null-space solutions, causing the robot's body to bend away from the obstacles while the tip continues to track the planned path. Occasionally, for obstacle avoidance, the tip may not align perfectly with planned points, resulting in a certain distance between them.

The second part critically examines the CDSR's dynamic motion planning capabilities. The simulation retains spatial dimensions consistent with Simulation 1. Two dynamic spherical obstacles are positioned similarly to the first part, with radii governed by $(60+10\sin{(\alpha t)})$mm and $(45+15\cos{(\beta t)})$mm, where $\alpha,\beta\in\mathbb{R}$. The trajectory planning spans from $[-50, 100, 390]^T$ to $[10, -120, 170]^T$. The damping coefficient is set at 0.3, and the obstacle-free coefficient is set at 40. For planning efficiency, collision assessments are streamlined to ten points along the robot's structure.

The dynamic planning procedure is shown in Fig. \ref{fig:simulation2}(a) and Fig. \ref{fig:simulation2}(b). Fig. \ref{fig:simulation2}(a-1) to (a-4) and (b-1) to (b-4) show the whole planning process with and without the collision-free planner, respectively. Fig. \ref{fig:simulation2}(a-5) and Fig. \ref{fig:simulation2}(b-5) show the minimum distance between the robot and the obstacles. At initial time step $t_0$, the method is invoked to derive a collision-free trajectory between the designated start and end points. This trajectory is further refined using spline interpolation, enabling the robot's tip to trace the proximate path point, thereby progressing to the subsequent time step, $t_1$. Due to fluctuating obstacle dimensions, the robot periodically updates its path based on its current position. This adjustment continues until it reaches the target. Notably, the robot can adjust to avoid collision with obstacles based on dynamic changes. The observed mean planning duration for each iteration stands at 4.64s, with a positional error of 0.47mm at the robot's tip. The collision-free planner exhibits better performance in maintaining safety.

\subsection{Dynamic Obstacle-Rich Motion Planning Comparison}
In this simulation, the goal is for each method to produce a sequence of points that guide the robot’s tip from the starting point to the goal point, along with the first three configurations. If the planned path or the robotic shape collides with an obstacle at any point, this planning will be marked as a failure. For every count of obstacles, 100 test scenarios were created. Each method underwent evaluation across these cases, where metrics included average running time (in seconds), success rate, and the average length (in millimeters) of successful trajectories. Time is discretized for simplicity into steps of 1 second each. Obstacles, modeled as spheres, have radii controlled by the function $(A+B\sin{(t)})$mm, with $A$ ranging from 30-50mm and $B$ from 10-20mm. These obstacles move linearly between the start and target points, covering a trajectory length of 50-100mm, influenced by $\sin({\cdot})$ or $\cos({\cdot})$ functions, keeping an average speed under 15mm/s. Given their speed, the motion capture system ensures obstacle tracking without data loss from large inter-frame intervals.

\vspace{-0.5em}
\begin{figure}[h]
    \centering
    \includegraphics[width=0.47\textwidth]{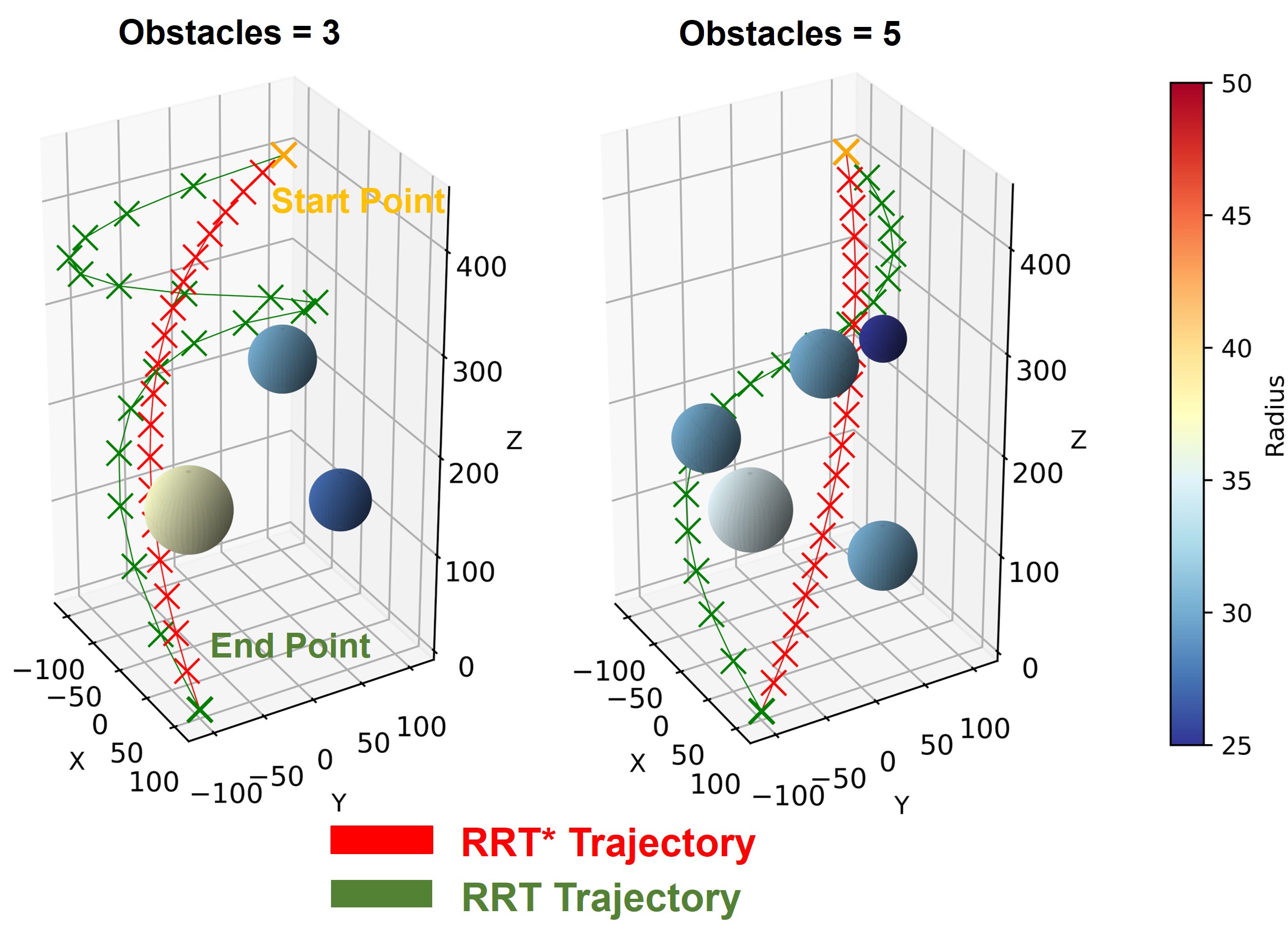}
    \caption{Comparison of trajectories generated by RRT and RRT* in multiple dynamic obstacles environments.}
    \label{fig:RRT_Compare}
\end{figure}
\vspace{-0.5em}

Table \ref{tab:table1} summarizes the key metrics of the simulation results. Within this simulated environment, we focused on comparing the performance of two methods: RRT-based (refer to as RRT) and RRT*-based (refer to as RRT*), for simplicity. The former algorithm disregards rewiring, while the latter incorporates it. Compared to \cite{2021Meng}, the scenario we proposed is more compact. It features larger obstacle dimensions and a denser environment. For the result, our proposed method significantly outperformed the techniques discussed in \cite{2021Meng}, reducing computational time by approximately 90\% and achieving a higher success rate. Although the sample-based motion planning method does not directly account for predicting the movement of obstacles, we fully consider the positions and sizes of obstacles at the start and end of each computational step during the path-checking process.

\begin{table}[h]
    \caption{SIMULATIONS OF DIFFERENT OBSTACLES}
    \label{tab:table1}
    \centering
    \resizebox{\columnwidth}{!}{
    \renewcommand{\arraystretch}{1.2} % adjust dis between word and line
    \begin{tabular}{|c|c|c|c|c|c|c|c|c|c|c|}
        \hline
            \multicolumn{3}{|c|}{\diagbox{Method}{\# Obs}} & \multicolumn{2}{c|}{1} & \multicolumn{2}{|c|}{3} & \multicolumn{2}{c|}{5} & \multicolumn{2}{c|}{7} \\
        \hline
        \multicolumn{3}{|c|}{RRT Time(s)/Succ} &3.05 &99\%	&4.77	&97\% &5.55 &95\%	&6.36	&90\%	\\ 
        \hline
        \multicolumn{3}{|c|}{RRT* Time(s)/Succ} &3.66 &99\%	&4.82	&89\% &6.07 &78\%	&7.30	&65\%	\\ 
        \hline
        \multicolumn{3}{|c|}{RRT Avg dis(mm)} & \multicolumn{2}{|c|}{715.36} & \multicolumn{2}{|c|}{736.11} & \multicolumn{2}{|c|}{785.46} & \multicolumn{2}{|c|}{743.36} \\
        \hline
        \multicolumn{3}{|c|}{RRT* Avg dis(mm)} & \multicolumn{2}{|c|}{609.07} & \multicolumn{2}{|c|}{626.70} & \multicolumn{2}{|c|}{625.46} & \multicolumn{2}{|c|}{597.45} \\
        \hline
    \end{tabular}}
    \vspace{-0.8em}
\end{table}

This approach allows us to effectively assess the feasibility of the planned path. By leveraging our proposed robot control technique based on safety constraints, we can quickly compute collision-free robot poses, allowing robots to track trajectory points without collisions. From table \ref{tab:table1}, it's evident that as the number of obstacles increases, the performance of both techniques diminishes. The introduction of the rewiring step in RRT* results in a longer runtime than RRT. Although RRT boasts a higher success rate, paths it generates are on average 17\% to 25\% longer than those by RRT*, suggesting the interpolated RRT* trajectory is smoother than the RRT trajectory (as shown in Fig. \ref{fig:RRT_Compare}). In an environment with dynamic obstacles, robots typically focus on tracking the initial 1-2 trajectory points. Compared to RRT, RRT* provides a superior steering direction for robot trajectory tracking. Therefore, though RRT has a high success rate, its unoptimized path requires a longer tracking time for the robot to move from the starting point to the endpoint.

\section{Real-World Experiment}
This section evaluates the performance of the RRT*-based safety-constrained method using the two-segment CDSR prototype in a real environment. Three servo motors control each segment of the robot. Employing five $\text{NOKOV}^\circledR$ motion capture cameras with 0.3mm accuracy, we instrumented markers to track the robot’s base, the proximal segment’s tip, and the distal segment’s end (see Fig. \ref{fig:Intro}). We recorded the robot's position and the associated actuation as it navigated the predefined path. Those recorded data revealed discrepancies between the robot's actual and predicted positions, providing insights into its configuration. The average planning time is approximately 8.06s. Considering the hardware limits, the algorithm meets the requirements for real-time planning.

\begin{figure}[h]
    \centering
    \includegraphics[width=0.45\textwidth]{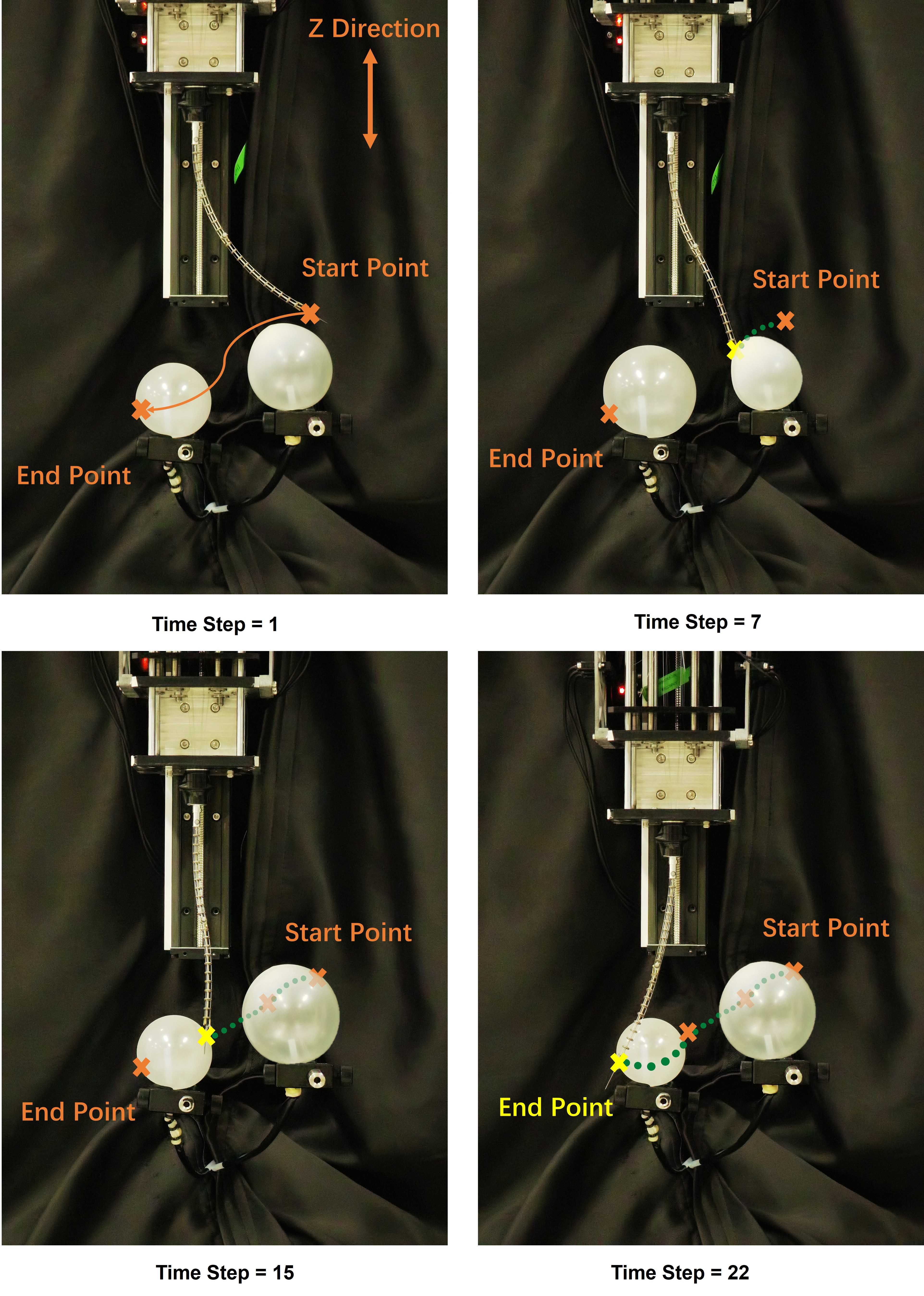}
    \caption{Real-world experiments on motion planning and trajectory tracking of CDSR robots in an environment with dynamic balloons. Yellow intersections indicate the current position the tip has reached. Green dots represent the arrived path.}
    \label{fig:RealEXP}
\end{figure}

Our proposed approach demonstrated satisfactory performance on the prototype robot in experimental evaluations. We select the initial position as $[140, 180, 370]^T$ and the target position as $[160, 70, 320]^T$. Two dynamic balloons are placed at $[125, 105, 335]^T$ and $[175, 175, 360]^T$. Their radii, controlled by different signal functions, range from 20mm to 35mm, respectively. Through our method, the robot could swiftly compute a collision-free trajectory from the initial position to the target position and successfully navigate around the obstacle. Fig. \ref{fig:RealEXP} illustrates the robot's collision-free path from start to end. The tip tracking error between the planned position and the actual arrival position is approximately 5.48mm. Our method guarantees secure and rapid responsive trajectory generation. We showcase its robustness in demanding scenarios, emphasizing security. This dual emphasis enhances its value for practical applications.

It should be noted that considering proximity to the target and obstacle avoidance during the planning process may result in slight deviations between the expected trajectory points and the actual tracking points. The parameters $D_{sf}$ and $\mu$ influence these deviations. Overall, our algorithm performs well in overcoming dynamic obstacles and efficiently accomplishing tasks.

\section{Conclusion And Future Work}
In this work, we developed a novel motion planning method for continuum robots operating in dynamic environments. This method builds upon the motion planning technique of RRT*, integrating dynamic obstacle safety constraints tailored for soft robots. The goal is to achieve collision-free planning for continuum robots in dynamic settings. Through simulations and experiments, we have demonstrated the method's effectiveness in handling complex dynamic environments. Future work will focus on refining avoidance strategies in dynamic environments, enhancing algorithm adaptability and robustness to handle unpredictably shaped obstacles, and optimizing algorithms to expedite planning and task execution. These advancements will further enable continuum robots to navigate dynamic environments.

\clearpage
\bibliographystyle{references/IEEE}
\bibliography{references/RRT.bib}

\end{document}